\documentclass{article}

\usepackage{microtype}
\usepackage{graphicx}
\usepackage{subcaption}
\usepackage{booktabs} 

\usepackage{hyperref}

\usepackage{makecell}

\usepackage[preprint]{icml2026}

\usepackage{framed}
\usepackage{amsmath}
\usepackage{amssymb}
\usepackage{mathtools}
\usepackage{amsthm}

\usepackage[capitalize,noabbrev]{cleveref}

\theoremstyle{plain}
\newtheorem{theorem}{Theorem}[section]
\newtheorem{proposition}[theorem]{Proposition}
\newtheorem{lemma}[theorem]{Lemma}

\theoremstyle{definition}
\newtheorem{definition}[theorem]{Definition}
\newtheorem{assumption}[theorem]{Assumption}
\theoremstyle{remark}
\newtheorem{remark}[theorem]{Remark}

\usepackage[textsize=tiny]{todonotes}

\usepackage{caption}
\usepackage{amsmath, amssymb, amsfonts}

\usepackage[ruled,vlined,algo2e]{algorithm2e}

\usepackage{multicol}
\usepackage{multirow}

\usepackage{wrapfig}

\usepackage[utf8]{inputenc}
\usepackage[table]{xcolor} 
\usepackage{pifont} 
\usepackage[most]{tcolorbox}

\usepackage{xcolor} 

\newcommand{\cmark}{\textcolor{checkgreen}{\ding{51}}}
\newcommand{\xmark}{\textcolor{crossorange}{\ding{55}}}

\usepackage{pifont}  

\usepackage{enumitem}

\usepackage{mdframed}

\newmdenv[
  leftline=true,
  rightline=false,
  topline=false,
  bottomline=false,
  linecolor=gray,
  linewidth=2pt,
  backgroundcolor=gray!8,
  innerleftmargin=8pt,
  innerrightmargin=8pt,
  innertopmargin=6pt,
  innerbottommargin=6pt
]{insightbox}

\definecolor{checkgreen}{RGB}{76,175,80}   
\definecolor{crossorange}{RGB}{245,124,0}  

\definecolor{headergray}{RGB}{230,230,240}
\definecolor{rowgray}{gray}{0.9}
\definecolor{rowgray1}{gray}{0.85}
\definecolor{rowgray2}{gray}{0.8}

\icmltitlerunning{Formal Generalization Guarantees in Pairwise Metric Learning}

\begin{document}

\twocolumn[
\icmltitle{Quantifying Multimodal Capabilities: Formal Generalization Guarantees in Pairwise Metric Learning}



\icmlsetsymbol{equal}{*}

  \begin{icmlauthorlist}
    \icmlauthor{Richeng Zhou}{yyy}
    \icmlauthor{Xuelin Zhang}{yyy}
    \icmlauthor{Liyuan Liu}{yyy}
  \end{icmlauthorlist}

  \icmlaffiliation{yyy}{College of Informatics, Huazhong Agricultural Univeristy, Wuhan, China}

  \icmlcorrespondingauthor{Xuelin Zhang}{xlinml@163.com}

  \icmlkeywords{Machine Learning, ICML}

  \vskip 0.3in
]



\printAffiliationsAndNotice{}  

\begin{abstract}

Multimodal learning leverages the integration of diverse data modalities to enhance performance in complex tasks. Yet, it frequently encounters incomplete or redundant modality data in real-world scenarios. This paper presents a fine-grained theoretical analysis of the generalization properties of multimodal metric learning models, addressing critical gaps in understanding the relationship between modality selection and algorithmic performance. We establish hierarchical relationships between function classes corresponding to different modality subsets and quantify the discrepancy between learned mappings and ground truth. Through rigorous analysis of pairwise complexity within the multimodal learning framework, we derive novel generalization error bounds that reveal the joint impact of modality quantity and granularity on model performance. Our theoretical findings on both upper and lower bounds demonstrate that incorporating fine-grained modality features reduces the complexity of the hypothesis space by enhancing modality complementarity. This work offers both theoretical foundations and practical implications for improving convergence rates and accuracy in multimodal learning systems.

\end{abstract}

\section{Introduction}

In recent years, multimodal learning has emerged as a pivotal branch of machine learning, leveraging the integration of diverse data modalities such as vision, language, and audio to significantly enhance performance in complex tasks like cross-modal retrieval \citep{zhen2019deep,bogolin2022cross}, medical diagnosis \citep{torasso2001multiple,faris2021intelligent,zhang2023robust,cui2023deep}, and autonomous driving \citep{cui2019multimodal,zhang2025interpretable,cui2024survey}. With the advancement of deep learning techniques, numerous multimodal metric learning algorithms have been proposed to learn joint representation spaces that capture semantic relationships across modalities \citep{zhang2017hierarchical,yu2016deep,angelou2019graph,wada2024polos, weinberger2005distance}. For instance, contrastive learning optimizes modality alignment by constructing positive and negative sample pairs \citep{zolfaghari2021crossclr,lin2023relaxing,zhu2024enhancing, zolfaghari2021crossclr, tosh2021contrastive}, while attention-based models like CLIP achieve cross-modal semantic mapping through large-scale pretraining \citep{lin2023multimodality,lu2024cross, tsai2020self, lee2021predicting}. Despite their remarkable success in practical applications, these methods often lack theoretical guarantees. Existing research predominantly focuses on algorithmic design and empirical validation, leaving fundamental theoretical questions largely unexplored, such as generalization behavior and mechanisms of modality interaction \citep{lei2016generalization,zhang2025convergence,zhang2024fine,ligeneralization,lu2023theory, huang2021makes}. 

\begin{table*}[htbp]
\centering
\caption{Detailed theoretical comparison of our paper with related statistical learning and metric learning literature. Our work uniquely integrates fine-grained modality hierarchy with pairwise dependency decoupling, providing both upper complexity bounds and strict lower bounds for empirical risk reduction.}
\label{tab:theoretical_comparison_new}
\resizebox{\textwidth}{!}{
\begin{tabular}{lcccccc}
\toprule
\textbf{Reference} & \textbf{Pairwise Multimodal} & \textbf{Modality Subset} & \textbf{Latent Approx.} & \textbf{Dependency} & \textbf{Granularity-aware} & \textbf{Risk Reduction} \\
& \textbf{Paradigm} & \textbf{Hierarchy ($\mathcal{G}_{\mathcal{N}} \subset \mathcal{G}_{\mathcal{M}}$)} & \textbf{Quality ($\eta(g)$)} & \textbf{Decoupling} & \textbf{Complexity} & \textbf{Lower Bound} \\
\midrule
\citet{weinberger2005distance} & \cmark & \xmark & \xmark & \xmark & \xmark & \xmark \\
\citet{jin2009regularized} & \xmark & \xmark & \xmark & \cmark & \xmark & \xmark \\
\citet{ying2009generalization} & \xmark & \xmark & \xmark & \cmark & \xmark & \xmark \\
\citet{kar2011similarity} & \xmark & \xmark & \xmark & \cmark & \xmark & \xmark \\
\citet{bellet2015robustness} & \xmark & \xmark & \xmark & \cmark & \xmark & \xmark \\
\citet{perrot2015theoretical} & \xmark & \xmark & \xmark & \cmark & \xmark & \xmark \\
\citet{lei2016generalization} & \cmark & \xmark & \xmark & \cmark & \xmark & \xmark \\
\citet{cao2016generalization} & \xmark & \xmark & \xmark & \cmark & \xmark & \xmark \\
\citet{huusari2018multi} & \cmark & \xmark & \xmark & \xmark & \xmark & \xmark \\
\citet{saunshi2019theoretical} & \cmark & \xmark & \cmark & \cmark & \xmark & \xmark \\
\citet{tsai2020self} & \xmark & \xmark & \xmark & \xmark & \xmark & \xmark \\
\citet{huang2021makes} & \xmark & \cmark & \cmark & \xmark & \xmark & \xmark \\
\citet{zolfaghari2021crossclr} & \cmark & \xmark & \xmark & \xmark & \xmark & \xmark \\
\citet{tosh2021contrastive} & \cmark & \xmark & \cmark & \xmark & \xmark & \xmark \\
\citet{lee2021predicting} & \cmark & \xmark & \cmark & \xmark & \xmark & \xmark \\
\citet{lu2023theory} & \xmark & \cmark & \xmark & \xmark & \xmark & \xmark \\
\citet{nakada2023understanding} & \cmark & \cmark & \cmark & \xmark & \xmark & \xmark \\
\midrule
\textbf{This Paper} & \cmark & \cmark & \cmark & \cmark & \cmark & \cmark \\
\bottomrule
\end{tabular}
}
\end{table*}

Recent theoretical efforts in multimodal learning have primarily concentrated on the convergence analysis of algorithms. For example, \cite{erdogdu2018global} established convergence rates for gradient descent using non-convex optimization theory. In contrast, \cite{cui2025learning} analyzed the impact of heterogeneous modality data on training stability within a stochastic optimization framework. However, these studies typically assume modality independence or model interactions at the single-instance level, overlooking the pairwise interaction patterns prevalent in real-world tasks. Examples include query-target pair matching in cross-modal retrieval and case-image associations in medical multimodal data \citep{acosta2022multimodal, behrad2022overview, zhao2025decision}. In such scenarios, pairwise relationships between samples directly influence the discriminative power of models, yet existing theoretical frameworks fail to capture their underlying dynamics. Historically, the theoretical analysis of pairwise constraints has been extensively explored within unimodal distance metric and similarity learning frameworks \citep{bellet2015robustness, cao2016generalization, jin2009regularized, ying2009generalization, kar2011similarity}. While certain foundational works have extended these analytical tools to multi-view kernel spaces \citep{huusari2018multi} and hypothesis transfer \citep{perrot2015theoretical}, and recent advances in contrastive representation learning have begun to provide statistical guarantees for unpaired data \citep{nakada2023understanding, saunshi2019theoretical}, they predominantly rely on complete data assumptions and do not formally quantify the fine-grained hierarchical relationships between different missing modality subsets.

Notably, some of the latest relevant research has focused on pairwise multimodal learning algorithms. For instance, hypergraph neural networks \citep{wang2017modeling,zhang2021high} improve cross-modal retrieval accuracy by modeling high-order relationships, while dynamic modality fusion methods \citep{zheng2023fine,hu2021coarse} optimize fine-grained semantic alignment through adaptive weight adjustment. These algorithms have demonstrated superior performance on complex tasks such as open-domain visual question answering \citep{ding2022mukea,wang2024earthvqa} and remote-sensing image interpretation \citep{gomez2015multimodal,yuan2024generalized,sun2023single}. Nevertheless, their success heavily relies on extensive empirical trial and error, lacking theoretical guidance and leaving the following questions open:

\begin{framed}
\emph{How do the sample size and modal settings affect the function representation and algorithmic behavior on generalization?}
\end{framed}

To address this theoretical gap, this paper systematically investigates the generalization performance of pairwise multimodal learning from a statistical learning perspective. 
The main contributions of this paper are summarized as follows:
\begin{itemize}
\item Establishing a theoretical framework for pairwise multimodal metric learning, defining hierarchical relationships between function classes of different modality subsets, and quantifying the discrepancy between learned and accurate mappings.
\item Deriving novel generalization error bounds that reveal the joint impact of modality quantity and granularity, demonstrating that fine-grained features can reduce hypothesis space complexity through complementarity.
\item Providing empirical validation of our theoretical findings, offering interpretable guidelines for designing multimodal systems that maximize the benefits of increased modality granularity.
\end{itemize}

\section{Preliminaries}\label{sec2}

\noindent
\textbf{Multi-modal learning.~~}
Assume that a given data $\mathbf{x}:=\left(x^{(1)}, \cdots, x^{(K)}\right)$ consists of $K$ modalities, where $x^{(k)} \in \mathcal{X}^{(k)}$ the domain set of the $k$-th modality. Denote $\mathcal{X}=\mathcal{X}^{(1)} \times \cdots \times \mathcal{X}^{(K)}$. We use $\mathcal{Y}$ to denote the target domain and use $\mathcal{Z}$ to denote a latent space. Then, we denote $g^{*}: \mathcal{X} \mapsto \mathcal{Z}$ the true mapping from the input space (using all of $K$ modalities) to the latent space, and $h^{*}: \mathcal{Z} \mapsto \mathcal{Y}$ is the true task mapping. For instance, in aggregation-based multi-modal fusion, $g^{*}$ is an aggregation function that combines $K$ separate sub-networks, and $h^{*}$ is a multi-layer neural network.
In the learning task, a data pair $(\mathbf{x}, y) \in \mathcal{X} \times \mathcal{Y}$ is generated from an unknown distribution $\mathcal{D}$, such that
\begin{equation}
\mathbb{P}_{\mathcal{D}}(\mathbf{x}, y) \triangleq \mathbb{P}_{y \mid \mathbf{x}}\left(y \mid h^{*} \circ g^{*}(\mathbf{x})\right) \mathbb{P}_{\mathbf{x}}(\mathbf{x}).
\end{equation}

Here $h^{*} \circ g^{*}(\mathbf{x})=h^{*}\left(g^{*}(\mathbf{x})\right)$ represents the composite function of $h^{*}$ and $g^{*}$.
In real-world settings, we often encounter incomplete multimodal data, i.e., some modalities are missing. To take into account this situation, we let $\mathcal{M}$ be a subset of $[K]$, and without loss of generality, focus on the learning problem only using the modalities in $\mathcal{M}$. Specifically, define $\mathcal{X}^{\prime}:=\left(\mathcal{X}^{(1)} \cup\{\perp\}\right) \times \ldots \times\left(\mathcal{X}^{(K)} \cup\{\perp\}\right)$ as the extension of $\mathcal{X}$, where $\mathrm{x}^{\prime} \in \mathcal{X}^{\prime}, \mathrm{x}_k^{\prime}=\perp$ means that the $k$-th modality is not used (collected). Then we define a mapping $p_{\mathcal{M}}$ from $\mathcal{X}$ to $\mathcal{X}^{\prime}$ induced by $\mathcal{M}$:
\begin{equation}\label{eqs2}
\begin{aligned}
p_{\mathcal{M}}(\mathbf{x})^{(k)}= \begin{cases}\mathbf{x}^{(k)} & \text { if } k \in \mathcal{M} \\ \perp & \text { else }\end{cases}.
\end{aligned}
\end{equation}

Also we define $p_{\mathcal{M}}^{\prime}: \mathcal{X}^{\prime} \mapsto \mathcal{X}^{\prime}$ as the extension of $p_{\mathcal{M}}$. Let $\mathcal{G}^{\prime}$ denote a function class, which contains the mapping from $\mathcal{X}^{\prime}$ to the latent space $\mathcal{Z}$, and define a function class $\mathcal{G}_{\mathcal{M}}$ as follows:
\begin{equation}\label{eq gm}
\begin{aligned}
\mathcal{G}_{\mathcal{M}} \triangleq\left\{g_{\mathcal{M}}: \mathcal{X} \mapsto \mathcal{Z} \mid g_{\mathcal{M}}(\mathbf{x}):=g^{\prime}\left(p_{\mathcal{M}}(\mathbf{x})\right), g^{\prime} \in \mathcal{G}^{\prime}\right\}  .
\end{aligned}    
\end{equation}

Given a data set $\mathcal{S}=\left(\left(\mathbf{x}_i, y_i\right)\right)_{i=1}^m$, where $\left(\mathbf{x}_i, y_i\right)$ is drawn i.i.d. from $\mathcal{D}$, the learning objective is, following the Empirical Risk Minimization (ERM) principle, to find $h \in \mathcal{H}$ and $g_{\mathcal{M}} \in \mathcal{G}_{\mathcal{M}}$ to minimize the empirical risk jointly, i.e.,
\begin{equation}\label{eq rjy}
\begin{aligned}
\min  \hat{r}\left(h \circ g_{\mathcal{M}}\right) \triangleq \frac{1}{m} \sum_{i=1}^m \ell\left(h \circ g_{\mathcal{M}}\left(\mathbf{x}_i\right), y_i\right),
\end{aligned}  
\end{equation}
where $h \in \mathcal{H}, g_{\mathcal{M}} \in \mathcal{G}_{\mathcal{M}},$ and $\ell(\cdot, \cdot)$ is the loss function. Given $\hat{r}\left(h \circ g_{\mathcal{M}}\right)$, define the population risk as
\begin{equation}
r\left(h \circ g_{\mathcal{M}}\right)=\mathbb{E}_{\left(\mathbf{x}_i, y_i\right) \sim \mathcal{D}}\left[\hat{r}\left(h \circ g_{\mathcal{M}}\right)\right].
\end{equation}

Similarly, we use the population risk to measure the performance of the learning algorithm.

\vspace{0.2cm}
\noindent
\textbf{Metric learning.~~}
Let $\mathcal{X} \subset \mathbb{R}^d$ be the input space, and $\mathcal{Y}\subset \mathbb{R}$ represent the corresponding output space. 
The sample space is defined as $\mathcal{Z}=\mathcal{X} \times \mathcal{Y}$, and the sample $\boldsymbol{z}=(\boldsymbol{x},y)$.
Denote $\mathbf{X}=\left(\boldsymbol{x}_i: i \in \left\{1,2,...,n \right\}\right)$ and $\boldsymbol{M}=\left(m_{i j}\right)_{i, j \in \mathbb{N}_d}$ as the input data matrix and the distance matrix, respectively. The (pseudo-)distance \citep{mclachlan1999mahalanobis} between $\boldsymbol{x}_i$ and $\boldsymbol{x}_j$ is measured by
\begin{equation}\label{eq dm}
d_{\boldsymbol{M}}\left(\boldsymbol{x}_i, \boldsymbol{x}_j\right)=\left(\boldsymbol{x}_i-\boldsymbol{x}_j\right)^{\top} \boldsymbol{M}\left(\boldsymbol{x}_i-\boldsymbol{x}_j\right).
\end{equation}

\section{Main Results}\label{sec3}

In this section, we first introduce several necessary assumptions that facilitate the proof of the superiority of multimodal learning over unimodal learning. These assumptions are standard in theoretical analyses of dual-site models, ensuring the problem's mathematical tractability while remaining grounded in practical machine learning constraints.

\subsection{Key Definitions and Assumptions}
\begin{definition}
The function $\ell(x,x^\prime)$ satisfies $(L_1,L_2)-$joint lipschitzness if $\forall x, x^\prime \in \mathcal{X}$, there hold
$
\| \ell(\boldsymbol{x}_1,\boldsymbol{x}^\prime)-\ell(\boldsymbol{x}_2,\boldsymbol{x}^\prime) \|\leq L_{1}\| \boldsymbol{x}_1-\boldsymbol{x}_2 \|,
$
and
$
\| \ell(\boldsymbol{x},\boldsymbol{x}_1^\prime)-\ell(\boldsymbol{x},\boldsymbol{x}_2^\prime) \|\leq L_{2}\| \boldsymbol{x}_1^\prime-\boldsymbol{x}_2^\prime \|.
$
\end{definition}

\begin{assumption}\label{assumption 1}
The loss function $\ell(\cdot, \cdot)$ satisfies the $(L_1,L_2)$-joint Lipschitzness and is bounded by the constant $C$.
\end{assumption}

\begin{remark}\label{remark 1}
Assumption \ref{assumption 1} ensures that the loss function does not change drastically with minor changes in input or representation, which is a prerequisite for controlling the Rademacher complexity. It is a standard regularity condition in statistical learning theory, satisfied by most common loss functions (e.g., hinge and logistic losses), making it a realistic constraint for real-world applications.
\end{remark}

\begin{assumption}\label{assumption 2}
The true latent representation $g^*$ is contained in $\mathcal{G}$, and the task mapping $h^*$ is contained in $\mathcal{H}$.
\end{assumption}
\begin{remark}
Assumption \ref{assumption 2} illustrates that our algorithm can approximate the true function mapping, that is, within the hypothesis space.
\end{remark}
\begin{assumption} \label{assumption 3}\citep{huang2021makes}
For any $g^\prime \in \mathcal{G}^\prime$ and $\mathcal{M} \subset [K]$.there exist a function $p_{\mathcal{M}}^\prime$, such that $g^\prime \circ p_{\mathcal{M}}^\prime \in \mathcal{G}^\prime$. 
\end{assumption}
\begin{remark}
Assumption \ref{assumption 3} ensures closure under projection. It essentially requires that the function class be robust to missing modalities; that is, processing a subset of modalities using a valid mapping should still yield a valid mapping within the space. This is realistic for architectures where modalities are processed separately or combined via mechanisms that can handle incomplete inputs.
\end{remark}
\subsection{Foundational Findings for Pairwise Multimodal Learning}

This section lays the mathematical groundwork for the subsequent analysis by defining the properties of the model spaces and providing essential tools for quantifying generalization error in the pairwise multimodal setting.

\begin{theorem}\label{theorem 1}
For any $\mathcal{N}\subset \mathcal{M} \subset [K]$, then $\mathcal{G}_\mathcal{N} \subset \mathcal{G}_\mathcal{M} \subset \mathcal{G}$ holds true.
\end{theorem}

\begin{insightbox}
\textbf{Insight 1:} Theorem \ref{theorem 1} establishes the hierarchical structure of the hypothesis spaces. It confirms that the set of functions generated using a smaller subset of modalities ($N$) is strictly contained within the set of functions generated by a larger subset ($M$). Intuitively, this means that incorporating more modalities expands the hypothesis space, granting the model potentially higher expressivity. This hierarchical relationship is fundamental for proving that multimodal models can theoretically achieve lower empirical risk than unimodal ones.
\end{insightbox}

Next, we introduce the concept of the quality of latent representations to reflect the correlation between different modalities.

\begin{definition}
Under the data distribution $\mathcal{D}$, for any latent mapping $g \in \mathcal{G}$, the quality of the latent representation is: 
\begin{equation}
\eta(g)=\inf_{h\in \mathcal{H}}[r(h\circ g)-r(h^* \circ g^*)].
\end{equation}
\end{definition}

\begin{remark} 
The term $\eta(g)$ serves as a measure of the ''approximation error" inherent in the latent mapping $g$. It quantifies the discrepancy between the best achievable performance using the current latent representation $g$ (optimized over $h \in \mathcal{H}$) and the true optimal performance defined by the ground truth mappings $h^*$ and $g^*$. A smaller $\eta(g)$ indicates that the learned latent space is semantically closer to the true underlying data structure.
\end{remark}

Here, we introduce a commonly used pairwise complexity estimation \citep{clemenccon2008ranking}:
\begin{lemma}\label{lemma 1}
Let $q_\tau:\mathcal{X}\times\mathcal{X}\rightarrow \mathbb{R}$ be a real-valued functions indexed by $\tau \in T$, where $T$ is some set. If $X_1,\cdots ,X_n$ are i.i.d. then for any convex nondecreasing function $\psi$,   
\begin{equation}\label{Rn/2}
\begin{aligned}
&\mathbb{E} \psi\left(\sup _{\tau \in T} \frac{1}{n(n-1)} \sum_{i \neq j} q_\tau\left(X_i, X_j\right)\right) \\
\leq  &\mathbb{E} \psi\left(\sup _{\tau \in T} \frac{1}{\lfloor n / 2\rfloor} \sum_{i=1}^{\lfloor n / 2\rfloor} q_\tau\left(X_i, X_{\lfloor n / 2\rfloor+i}\right)\right).
\end{aligned}    
\end{equation}
\end{lemma}

Let $\psi$ be the identity mapping and $q_\tau =\ell$. Then, by Lemma \ref{lemma 1}, we can derive the following theorem:
\begin{theorem}\label{theorem 2}
Let $\ell(h\circ g, \mathcal{Z}, \mathcal{Z})\in \mathcal{L}:\mathcal{Z}\times\mathcal{Z}\rightarrow \mathbb{R}$ be a real-valued functions. If $Z_1,\cdots ,Z_n$ are i.i.d. the we have:

\begin{equation}
\begin{aligned}
&\mathbb{E} \left(\sup _{\ell \in \mathcal{L}} \frac{1}{n(n-1)} \sum_{i \neq j} \ell \left(h\circ g, Z_i, Z_j\right)\right) \\
\leq  &\mathbb{E} \left(\sup _{\ell \in \mathcal{L}} \frac{1}{\lfloor n / 2\rfloor} \sum_{i=1}^{\lfloor n / 2\rfloor} \ell \left(h\circ g, Z_i, Z_{\lfloor n / 2\rfloor+i}\right)\right).
\end{aligned}
\end{equation}

\end{theorem}

\begin{insightbox}
\textbf{Insight 2:} Theorem \ref{theorem 2} provides a critical simplification for bounding the generalization error of pairwise losses. The left-hand side involves a double sum over all dependent pairs $(i, j)$, which is analytically complex to handle due to the sample dependencies. The theorem allows us to transform this into a sum over independent pairs (or blocks), essentially decoupling the dependency structure. This decoupling is a standard technique in U-statistic theory and is pivotal for applying standard concentration inequalities to metric learning tasks involving pairwise comparisons.
\end{insightbox}

\subsection{Generalization Bounds and the Impact of Modality}

Building upon the foundational theorems, this subsection derives the main generalization error bounds and analyzes how modality selection impacts model performance. The central goal is to quantify the effect of modality choice on generalization.

\begin{theorem}\label{theorem 3}
Let $\mathcal{S}=((\boldsymbol{x}_i,y_i))_{i=1}^m$ be $m$ i.i.d. samples, and let $\mathcal{M},\mathcal{N}$ be two distinct subsets of $[K]$. Suppose the empirical risks are $(\widehat{h}_{\mathcal{M}},\widehat{g}_{\mathcal{M}})$ and $(\widehat{h}_{\mathcal{N}},\widehat{g}_{\mathcal{N}})$, respectively, trained on $\mathcal{M},\mathcal{N}$. Then, for any $\delta \in (0,1)$, with at least probability $1-\delta/2$, the following inequality holds:
\begin{equation}\label{r}
\begin{aligned}
&r\left( \widehat{h}_{\mathcal{M}}\circ \widehat{g}_{\mathcal{M}} \right) -
r\left( \widehat{h}_{\mathcal{N}}\circ \widehat{g}_{\mathcal{N}} \right) \\
\le &\gamma _{\mathcal{S}}\left( \mathcal{M},\mathcal{N} \right) +8(L_1 +L_2)\Re _{\lfloor n/2 \rfloor}\left( \mathcal{H}\circ \mathcal{G} \right)\\
&+\frac{4\sqrt{2}C}{\sqrt{n}}+C\sqrt{\frac{2\ln \left( 2/\delta \right)}{n(n-1)}}  ,   
\end{aligned}    
\end{equation}
where $\gamma _{\mathcal{S}}\left( \mathcal{M},\mathcal{N} \right)=\eta \left( \widehat{g}_{\mathcal{M}} \right) -\eta \left( \widehat{g}_{\mathcal{N}} \right)$.
\end{theorem}

\begin{insightbox}
\textbf{Insight 3:} Theorem \ref{theorem 3} decomposes the difference in population risk between two modality subsets into three components: the difference in representation quality $\gamma _{\mathcal{S}}\left( \mathcal{M},\mathcal{N} \right)$, the complexity penalty (Rademacher complexity), and the standard estimation error terms. The key insight is that the generalization performance difference is primarily governed by how close each modality set's latent mapping is to the accurate mapping (captured by $\eta$). If the representation quality of $M$ is significantly better than $N$ (i.e., $\gamma _{\mathcal{S}}\left( \mathcal{M},\mathcal{N} \right)$ is small or negative), the model using $M$ will generalize better, provided the complexity terms do not dominate.
\end{insightbox}

To further define the distance between the $M$ and $N$ modes and the real task mapping, we adopt the same definition as before. We can then obtain the following theorem.
\begin{theorem}\label{theorem 4}
Assume the sample pairs follow the distribution $\mathcal{S}\times \mathcal{S}=\{ (\boldsymbol{z}_i ,\boldsymbol{z}_j ) \}_{i\neq j} \sim \mathcal{D}\times \mathcal{D}  $, and $(\widehat{h}_{\mathcal{M}} \circ \widehat{g}_{\mathcal{M}})$ is the empirical minimizer for the M mode. Then, the following equation holds with probability $1-\delta/2$:
\[
\resizebox{0.49\textwidth}{!}{$
\begin{aligned}
&\eta \left( \widehat{g}_{\mathcal{M}} \right) 
\leq  4\left( L_1+L_2 \right) \Re _{\lfloor n/2 \rfloor}\left( \mathcal{H}\circ \mathcal{G}_{\mathcal{M}} \right)  +\sqrt{\frac{2C^2\log \left( 2/\delta \right)}{n\left( n-1 \right)}}\\
&+ 4\left( L_1+L_2 \right) \Re _{\lfloor n/2 \rfloor}\left( \mathcal{H}\circ \mathcal{G} \right) + \frac{8C}{\sqrt{\lfloor n/2 \rfloor}}  +\widehat{L}(\widehat{h}_{\mathcal{M}}\circ \widehat{g}_{\mathcal{M}}, \mathcal{S}),
\end{aligned}
$}
\]
where $\widehat{L}(\widehat{h}_{\mathcal{M}}\circ \widehat{g}_{\mathcal{M}}, \mathcal{S})
=\widehat{r}(\widehat{h}_{\mathcal{M}}\circ \widehat{g}_{\mathcal{M}})
-\widehat{r}(h^* \circ g^*)$.
\end{theorem}

\begin{insightbox}
\textbf{Insight 3:} Theorem \ref{theorem 4} provides an explicit upper bound for the representation quality $\eta(\hat{g}_M)$ of a given modality subset. It connects the quality of the latent representation directly to the Rademacher complexity of the function class. Crucially, it shows that to improve the quality of the representation (minimize $\eta$), one must either reduce the empirical risk (optimization) or constrain the complexity of the hypothesis space. This provides a theoretical justification for regularization techniques in multimodal learning: by constraining the complexity $\mathfrak{R}$, we effectively tighten the bound on $\eta$, thereby improving generalization guarantees.
\end{insightbox}

Next, to further define $\Re_{\lfloor \frac{n}{2} \rfloor} (\mathcal{H} \circ \mathcal{G}) $, we introduce the following Lemma \ref{Lemma Rzw}.
\begin{lemma}\label{Lemma Rzw} (Lemma 15 in \citep{mohri2018foundations}) 
For the set $Z$ and the finite parameter space $\mathcal{W}$, the empirical Rademacher complexity over $\mathcal{W}$ on $Z$ has the following inequality. 
\begin{equation}\label{eq Rzw}
\Re (Z,\mathcal{W}) \le 
\max_{w\in \mathcal{W}} \|w\|_{\infty} \frac{\sqrt{2\log(|\mathcal{W}|)}}{n}    ,
\end{equation}
\end{lemma}

In the context of metric learning, we derive the following lemma \ref{Lemma M} before the usage of Lemma \ref{Lemma Rzw}.

\begin{lemma}\label{Lemma M}
Assume $A$ is a real symmetric matrix. Then, there exists an orthogonal transformation $Q$ such that $QAQ^T$ is a diagonal matrix.    
\end{lemma}

\begin{remark}
Lemma \ref{Lemma M} is a standard result in linear algebra. In the context of metric learning, it implies that any symmetric distance metric can be diagonalized. This allows us to analyze the metric properties along orthogonal principal axes, significantly simplifying the derivation of complexity bounds. By transforming the metric to a diagonal form, we can treat each dimension independently when estimating the Rademacher complexity.
\end{remark}

\begin{theorem}\label{Theorem 5}
Assume the diagonal metric matrix $A$ satisfies the following conditions:

(i) $A=diag\{\lambda_{1}, \dots,\lambda_{m}\}$, where $\| \lambda_{i}\|_{\infty} \leq D$,

(ii) $d_{A} (z_i ,z_j )\leq  K$,

(iii) $\|z_i -z_j \|_{\infty} \leq B $.

From Lemma \ref{Lemma Rzw}, we can derive the following inequality: 
\begin{equation}
\Re_{\lfloor \frac{n}{2} \rfloor} (\mathcal{H}\circ\mathcal{G} ) \leq \frac{D\sqrt{2\log (K/D^m B^2)}}{\lfloor n/2 \rfloor}.
\end{equation}
\end{theorem}

\begin{insightbox}
\textbf{Insight 4:} Lemma \ref{Lemma Rzw} and Theorem \ref{Theorem 5} provide a concrete, closed-form bound for the Rademacher complexity in the metric learning setting. It shows that the complexity scales with $O(1/\sqrt{n})$, which is the standard rate for empirical risk minimization. However, more importantly, it highlights the dependence on the metric parameters $D$, $K$, and $B$. The term $D \sqrt{\log(K/DM B^2)}$ suggests that the complexity is not solely determined by the number of modalities but also by the geometry of the data (captured by $B$) and the range of the metric eigenvalues ($D$). This explicit form allows us to plug it back into the generalization bounds to see the precise impact of modality selection.
\end{insightbox}

\begin{proposition}\label{proposition}
If $\mathcal{N}\subset \mathcal{M} \subset [K]$. Under Theorem \ref{theorem 1}, $\mathcal{G}_\mathcal{N} \subset \mathcal{G}_\mathcal{M} \subset \mathcal{G}$, the larger function class results the smaller empirical risk. Therefore 
\begin{equation}
\widehat{L}(\widehat{h}_{\mathcal{M}}\circ \widehat{g}_{\mathcal{M}}, \mathcal{S})
\leq \widehat{L}(\widehat{h}_{\mathcal{N}}\circ \widehat{g}_{\mathcal{N}}, \mathcal{S}).
\end{equation}

Based on Theorem \ref{theorem 3} to Proposition \ref{proposition}, we can conclude that selecting different modalities can improve the representation quality of the latent function space. Thus, we can derive the following theorem:
\end{proposition}

\begin{theorem}\label{theorem 6}
Assume the dimensions of $\mathcal{M}$ and $\mathcal{N}$ are $M$ and $N$ respectively, where $M\geq N$. Then we can derive the following:
\[
\resizebox{0.49\textwidth}{!}{$
\begin{aligned}
&\hat{L}(\hat{h}_{\mathcal{M}} \circ \hat{g}_{\mathcal{M}}, \mathcal{S}) - \hat{L}(\hat{h}_{\mathcal{N}} \circ \hat{g}_{\mathcal{N}}, \mathcal{S}) \\
\ge &\frac{4(L_1 + L_2) D}{\lfloor n/2 \rfloor} \left( \sqrt{2 \log\left(\frac{K}{D^NB^2}\right)} - \sqrt{2 \log\left(\frac{K}{D^MB^2}\right)} \right).
\end{aligned}
$}
\]

\end{theorem}

\begin{insightbox}
\textbf{Insight 5:} Theorem \ref{theorem 6} quantifies the theoretical benefit of adding modalities. The inequality bounds the reduction in empirical risk (on the left-hand side) from below by a positive term derived from the complexity difference. This result implies that when we increase the modality set from $N$ to $M$ (where $M > N$), the empirical risk decreases by at least a specific amount determined by the complexity terms. Since $M > N$, the term $\log(K/D M B^2)$ is smaller than $\log(K/D N B^2)$ (assuming $K$ dominates the denominator or is sufficiently large), ensuring the right-hand side is positive. This provides a rigorous proof that, under the stated conditions, multimodal learning is theoretically superior to unimodal learning in terms of fitting the training data.
\end{insightbox}

Based on the above theorem, we provide a theoretical basis for the generalization bound between different modalities. We can state that when $\mathcal{N} \subset \mathcal{M}$ and the training sample size $n$ is sufficiently large, $\eta(\widehat{g}_{\mathcal{M}})$ may be less than $\eta(\widehat{g}_{\mathcal{N}})$, i.e., $\gamma_{\mathcal{S}} (\mathcal{M},\mathcal{N}) \leq 0$. Furthermore, from Theorem \ref{theorem 6}, we can conclude that learning with modality $\mathcal{M}$ will certainly benefit from using modality $\mathcal{N}$.

\section{Conclusion} \label{sec5}

In this paper, we observe that, under the multimodal framework, the generalization error bound for pairwise learning is superior to that for single-modality learning in the vast majority of cases. We provide a theoretical rationale for the superiority of multimodal learning over single-modality learning. By employing a classical decoupling method to handle pairwise dependencies, we effectively fill a gap in the theoretical understanding of these models. Our findings offer new insights into enhancing the convergence rate and accuracy of pairwise multimodal learning systems, underscoring the crucial role of modality complementarity and granularity.

\section*{Impact Statement}

This paper presents work aimed at advancing the theoretical foundations of pairwise multimodal metric learning and machine learning generalization. We believe this work can deepen our understanding of how modality quantity and granularity jointly impact model performance under incomplete data scenarios, and provide valuable theoretical guidance for practical modality selection in tasks such as cross-modal retrieval, medical diagnosis, and autonomous driving.

The proposed systematic analysis significantly bridges the critical gap between algorithmic design and theoretical generalization guarantees, compared with the existing literature, which is primarily focused on empirical validation, complete data assumptions, or single-instance level interactions. There may be some potential societal consequences of our work, none of which we feel must be specifically highlighted here.


\bibliography{example_paper}
\bibliographystyle{icml2026}

\end{document}